\documentclass[journal]{IEEEtran}
\usepackage{cite}

\ifCLASSINFOpdf
  \usepackage[pdftex]{graphicx}
\else
  \usepackage[dvips]{graphicx}
\fi
\begin{document}
%
\title{How an Electrical Engineer Became an Artificial Intelligence Researcher, a Multiphase Active Contours Analysis}
%
%
%

\author{Kush R.~Varshney
\thanks{K.\ R.\ Varshney is with IBM Research, Thomas J.\ Watson Research Center, Yorktown Heights,
NY, 10598 USA e-mail: krvarshn@us.ibm.com.}
}

\maketitle

\begin{abstract}
This essay examines how what is considered to be artificial intelligence (AI) has changed over time and come to intersect with the expertise of the author. Initially, AI developed on a separate trajectory, both topically and institutionally, from pattern recognition, neural information processing, decision and control systems, and allied topics by focusing on symbolic systems within computer science departments rather than on continuous systems in electrical engineering departments. The separate evolutions continued throughout the author's lifetime, with some crossover in reinforcement learning and graphical models, but were shocked into converging by the virality of deep learning, thus making an electrical engineer into an AI researcher.  Now that this convergence has happened, opportunity exists to pursue an agenda that combines learning and reasoning bridged by interpretable machine learning models.
\end{abstract}


%
\IEEEpeerreviewmaketitle

\section{Introduction}
%
%
%
%
\IEEEPARstart{H}{ow} is it that an electrical engineer versed in the theories and methods of signals and systems, detection and estimation, and pattern recognition now finds himself as a researcher in the artificial intelligence (AI) department of an industrial laboratory?  In this personal reflection, I follow the evolution of two active contours in an attempt to answer this question: (I) the contour of the field of AI itself, and (II) the contour of my own interests and expertise. The reader may keep in mind the analogy of active contours from the computer vision literature: gradient descent flows of boundary shapes to partition spaces \cite{ChanV2001,VeseC2002}.

\section{Setting}

It is the AI winter, 26 years since the conclusion of the mythical Dartmouth Summer Research Project on Artificial Intelligence.  One of the founders of AI who was present at Dartmouth, Allen Newell, says \cite{Newell1982}, ``By almost any account pattern recognition and AI should be a single field, whereas they are almost entirely distinct.''  One of the leading researchers in stochastic systems, Alan Willsky, says \cite{Willsky1982},
\begin{quotation}
``As I see it, there is a natural marriage here.  The perspective of AI is aimed directly at attacking and breaking down problems of enormous complexity into smaller problems.  On the other hand, the perspective in control, estimation, decision, and system theory is to solve very precisely specified (and usually small) problems and to provide the means for quantitative evaluation of performance for these solutions.  \ldots There might be significant payoff from a dialogue among researchers in AI and in decision and control.''
\end{quotation}
I am also born in this same season of Autumn 1982.

\section{Initialization I}

Although it was obvious to informed observers and participants in 1982 that there was no good reason for such a separation of AI from pattern recognition and decision systems, the separation had become entrenched in the decade following Dartmouth.  The field of artificial intelligence had origins in cybernetics \cite{Kline2011}, the study of control and communication in the animal and the machine, but was fully focused on a distinct set of topics in short order.  This set of topics was coherent along several different dimensions, notably with a focus on \cite{Newell1982}:
\begin{itemize}
	\item symbolic systems versus continuous systems;
	\item problem-solving versus recognition;
	\item psychology versus neuroscience;
	\item performance versus learning; and
	\item serial versus parallel.
\end{itemize}
The contrasting focus in each of the different dimensions was the purview of cybernetics and pattern recognition.  Not only was there a clear topical separation, but an institutional one as well with AI researchers in computer science departments and cyberneticians in electrical engineering departments, with nary a crossover \cite{Newell1982}.  

The group in electrical engineering departments connected with control theorists, information theorists, signal processors, statisticians and so on at established venues such as the Allerton Conference on Communication, Control, and Computing, the Conference on Information and Systems and Sciences, the Asilomar Conference on Signals, Systems, and Computers, and the Conference on Decision and Control but did not interact with the AI researchers.  In fact, ``at the second International Joint Conference on Artificial Intelligence in London, it was decided that henceforth the conference would not accept pure pattern-recognition papers''  \cite{Newell1982}.  At this time, new venues such as the International Conference on Pattern Recognition, the Conference on Computer Vision and Pattern Recognition, and the Conference on Neural Information Processing Systems (NIPS) were created separately from AI venues.

\section{Initialization II}

My own evolution has beginnings before I am born through my grandfather and father.  In the years immediately after Dartmouth, my grandfather came to the United States from India to study systems theory and control within the electrical engineering department of the University of Illinois at Urbana-Champaign.  By 1982, he had made contributions to sensitivity theory of linear systems and to the application of systems and control concepts to agriculture and economics \cite{VarshneyP1972,VarshneyP1974,Varshney1976}.  When I was born, my father was a young professor of electrical engineering at Syracuse University, having published on signal quantization and detection as well as information-theoretic construction of decision trees \cite{VarshneyH1978,Varshney1980,Varshney1981,HartmannVMG1982}.

\section{Contour Evolution I}

Mostly independent of core AI research, several approaches to supervised machine learning were proposed and further developed, including neural networks \cite{LeCunBDHHHJ1989}, kernel methods \cite{BoserGV1992}, and ensemble methods \cite{Ho1995}.  There were starts and stops along the way, but a seemingly steady course of development included advances in the theory and illustrative application to realms such as document analysis and data mining.  These developments tended to be exactly the ones that the International Joint Conference on Artificial Intelligence (IJCAI) was targeting to put outside the contour of AI, but were a growing lot over time.  From an institutional perspective, machine learning grew in computer science departments, but was still mostly a distinct endeavor from AI.  

Bernard Widrow had been one of the visitors to the Dartmouth meeting and later developed algorithms at the heart of adaptive filtering \cite{WidrowH1960}.  There were connections forged between adaptive signal processing, control theory, and backpropagation (the workhorse of neural network training today) \cite{NarendraP1990,HassibiSK1993}.  As described in \cite{SuttonB1998}, optimal control (especially approximate dynamic programming) was also one of the threads, along with trial-and-error approaches and temporal-difference methods \cite{Tesauro1995} to come together to form the main thrust of reinforcement learning, namely Q-learning \cite{WatkinsD1992}.  Reinforcement learning, as it developed, was still also more centered around continuous systems, neuroscience, and learning.  However, because of its usefulness in problem-solving it was more recognized to be within the contour of AI than pure supervised learning.

Probabilistic graphical models \cite{HintonS1983,Pearl1988}, a means to ``breaking down problems of enormous complexity into smaller problems,'' came to be a direction of significant interest and development in pattern recognition and machine learning as well as applied statistics. As evidence, note that \cite{Bishop2006} has a large chapter devoted to graphical models along with chapters on neural networks and kernel methods.  Moreover, graphical models provided a connection point for distributed detection and distributed control \cite{SandellVAS1978,ViswanathanV1997}, especially as sensor networks gained relevancy \cite{CetinCFIMWW2006}. However, graphical models were also inside the contour of core AI, especially belief networks and constraint satisfaction problems with their relevance to symbolic systems, problem-solving and psychology. 

\begin{figure}
\includegraphics[width=\columnwidth]{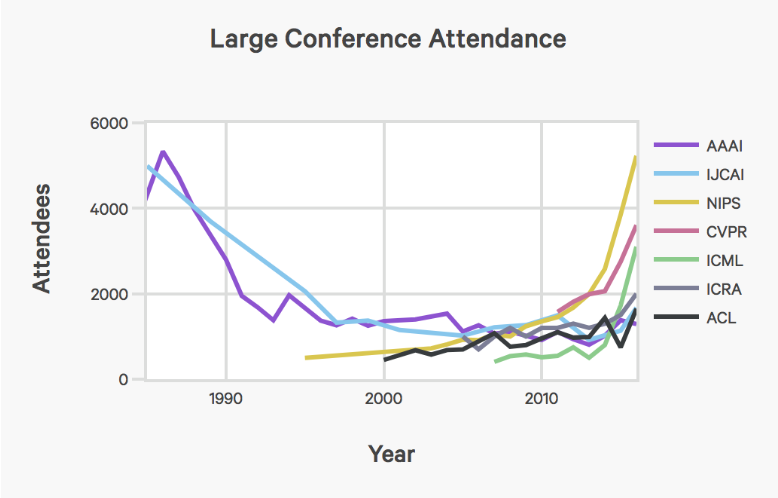}
\caption{Evolution of conference attendance \cite{ShohamPBC2017}.}
\label{fig:confattend}
\end{figure}

So-called good old-fashioned AI \cite{Haugeland1985}, the symbolic core inside the contour of AI, continued to progress independently of supervised learning ideas.  Expert systems built upon heuristic search, planning, logic programming, default reasoning, and knowledge representation led to impressive demonstrations such as the defeat of the world chess champion by an AI system \cite{Campbell1999} and wide adoption in industry.  There was a wane in research interest, as exemplified by the decrease in attendance at IJCAI and the Association for the Advancement of AI Conference on AI (AAAI) seen in Figure~\ref{fig:confattend}, taken from \cite{ShohamPBC2017}.  

The boundary of AI was fairly consistently viewed by both practitioners and outside observers.  Symbolic AI was AI, and continuous system machine learning was only said to be AI if push came to shove.  In industrial application, machine learning manifested as data mining, advanced predictive analytics, and eventually data science. There were some changes afoot though. For example, the Artificial Intelligence Laboratory merged with the Laboratory for Computer Science at the Massachusetts Institute of Technology.

The contour evolution changed dramatically with very large datasets becoming available in certain application domains (primarily recognition) and the very large accuracy improvements achieved in those domains by deep neural networks, helped along by the computation power of graphics processors \cite{LeCunBH2015}.   The virality of deep learning (cf.\ NIPS attendance in Figure~\ref{fig:confattend}) brought supervised machine learning into the limelight of outside observation and also quickly brought the label of AI onto it from the outside in.  Startups and established companies added `AI' to the ends of their names.  In a problem-solving task, reinforcement learning with deep neural networks achieved superhuman performance at Go \cite{Silver2017}.

The connection of Boltzmann machines to both graphical models and neural networks could have played a role in bringing neural networks inside the contour of AI from the practitioner perspective, but it does not appear to be the case.  AAAI had no accepted papers with the word `neural' in the title in 2013, 6 papers in 2014, 7 papers in 2015, 30 in 2016, 29 in 2017, and 92 in 2018, which indicates that the acceptance of non-symbolic, neural-oriented work into core AI was driven by the hype; otherwise the big increase would have started earlier.  

Now the contour evolution of AI appears to have completely subsumed neuro-inspired continuous parallel learning systems for recognition from both the observer and practitioner perspectives.  IJCAI 2018 is co-located with ICML under the banner of the Federated AI Meeting, which would have been unthinkable even a couple of years ago. The reunification of symbolic AI with the descendants of cybernetics and pattern recognition seems to be near-complete.

\section{Contour Evolution II}

Whether due to nature, nurture or both, my own evolution shows a proclivity towards signals and systems, detection and estimation, and pattern recognition and machine learning from the electrical engineering perspective.  

I was generally interested in and proficient at all subjects in school when I did a small research stint at the nuclear medicine department of the local university hospital during a high school summer vacation.  I loaded radioiodine images into IDL with the goal of estimating the volume of the thyroid gland using active contours \cite{KassWT1988}.  At that point, I only had a rough sense of what my grandfather and father worked on, so I did not appreciate that this was in the same vein.  I did not make headway because I had not yet learned calculus and computer programming, but the experience set the gradient flow direction of my personal evolution.  

In college, I majored in electrical and computer engineering and minored in computer science, taking many signal processing, probability, telecommunications, computer graphics, scientific computing and other continuous systems oriented classes.  A summer research project on classifying segments of postal images (images of envelopes, etc.) provided a practical introduction to pattern recognition and document analysis \cite{Varshney2004}.  A discrete math class was the furthest I got into the study of symbolic systems.  A computer architecture-oriented internship raised a barrier for my evolution in the direction of computer engineering.  

My masters thesis, in Alan Willsky's group, was on radar imaging from a sparse signal reconstruction perspective right before the hype of compressed sensing began \cite{VarshneyCFW2008}.  I did a medical imaging research stint during the summer after submitting my masters thesis, this time equipped with calculus and programming that allowed me to actually work on active contours methods successfully \cite{VarshneyPDKRHR2009}.  I was seemingly headed down the path of a career in image processing, but the evolution did not proceed in this way.  

I was surrounded by a lot of students working on graphical models research and I started looking into that direction.  In fact, I even made the arrangements for a talk by Hinton in April 2008 on ``An Efficient Way to Learn Deep Generative Models,'' which in retrospect, was an early warning of the impending wave of deep learning. However, my foray into graphical models did not last, as the research did not speak to me in the same way that the elegance of signal detection theory did.  A side project utilizing fundamental properties of Bayes optimal decision rules \cite{VarshneyV2008}, auditing a machine learning class in which I saw perceptrons and kernel methods presented formally for the first time, and a moment of inspiration that the standard supervised classification problem in general feature spaces could be approached using active contours methods led to the main research direction for my doctoral dissertation in supervised learning \cite{VarshneyW2010}.  An internship at a national laboratory on ensemble classifiers rounded out my machine learning experiences \cite{VarshneyPMCH2013}.  Throughout this whole evolution, I never thought of myself as an artificial intelligence researcher; I continued to think of myself as an electrical engineer, but one who had moved from signals to finite data.

As I was finishing my degree, I started reading about and getting excited about the use of data-driven and machine learning-driven decision making in applications I had not considered before \cite{Baker2008}.  I joined the mathematical sciences department at the IBM T.\ J.\ Watson Research Center to pursue the same.  I was doing applied predictive analytics research and publishing in data mining, signal processing, and machine learning venues: primarily supervised learning throughout \cite{VarshneyM2011,BaierCCLMRSSV2012,VarshneyCFWFM2014,VarshneyCANSXS2015,VarshneyV2016,CalmonWVRV2017,Varshney2017,WeiRV2018}.  Predictive analytics morphed into data science, but it still was not AI, it was industrial and applied mathematics of the continuous systems variety.  Some of the other researchers in the department had transitioned into analytics and data science from core AI and were not doing symbolic research anymore.  

Just recently, at the end of 2017, I became an AI researcher in the view of the IBM Research division, the overall IBM company, and the external world not because of any change in the research content I was pursuing (albeit with encouragement for more publishable research rather than industrial solutions), but because of a reorganization and the changing contour of AI itself that put machine learning squarely inside.  My evolution, although not as straightforward as can be, remained fully outside the contour of AI until the shock of deep learning changed both the contour of AI and the priorities of the company.  It is thus that an electrical engineer became a machine learning researcher who became an AI researcher.

\section{Interpretation}

So now what? We have seen the virality of deep learning create a shock that has brought symbolic AI and machine learning within a single contour just in the last year.  Can this rejoining after more than 50 years lead to something beyond what could be achieved in isolation?

Reprising \cite{Newell1982}, note that:
\begin{quotation}
``Adopting a class of systems has a profound influence on the course of a science. Alternative theories that are expressed within the same class are comparable in many ways. But theories expressed in different classes of systems are almost totally incomparable. Even more, the scientist's intuitions are tied strongly to the class of systems he adopts -- what is important what problems can be solved, what possibilities exist for theoretical extension, etc.''
\end{quotation}

Deep learning has seen the most success thus far in the presence of very large datasets, but most real-world applications do not afford such data.  Can the combination of machine learning and machine reasoning \cite{Bottou2014} help us extend the success to small data regimes?  Symbolic reasoning builds upon logic, but deep neural networks are black boxes whose models and outputs are difficult to reason about.  Rule set classifiers learned from data  \cite{Cohen1995,SuWVM2016,MalioutovVED2017}, on the other hand, are inherently presented in the form of logical expressions of features and can form the bridge between supervised learning and logical reasoning, and have the ancillary benefit of interpretability that many real-world applications (especially high stakes ones) require for safety \cite{VarshneyA2017}. 

The combination of learning and reasoning through the platform of directly interpretable models such as rule sets seems to be a promising direction to further the overarching goal of artificial intelligence.  The dimensions of distinction: symbolic vs.\ continuous, problem-solving vs.\ recognition, psychology vs. neuroscience, performance vs.\ learning, and serial vs. parallel all get bridged via such an approach.  The contour evolution of AI going forward may be transparent and bright if done right and not hindered by historic distinctions.


%



\section*{Acknowledgment}

The author thanks M.\ Campbell, M.\ \c{C}etin, B.\ Y.\ Chen, T.\ L.\ Fine, J.\ W.\ Fisher, III, S.\ S.\ Hemami, C.\ R.\ Johnson, Jr., A.\ Krol, A.\ Mojsilovi\'c, N.\ Paragios, T.\ W.\ Parks, R.\ K.\ Varshney, P.\ K.\ Varshney, and A.\ S.\ Willsky for shaping his contour evolution.

\ifCLASSOPTIONcaptionsoff
  \newpage
\fi



\bibliographystyle{IEEEtran}
\bibliography{cybernetics}
\end{document}